\documentclass[conference]{IEEEtran}
\IEEEoverridecommandlockouts
\usepackage{cite}
\usepackage{amsmath,amssymb,amsfonts}
\usepackage{algorithmic}
\usepackage{graphicx}
\usepackage{textcomp}
\usepackage{xcolor}
\usepackage{multirow}
\usepackage[normalem]{ulem}
\useunder{\uline}{\ul}{}
\def\BibTeX{{\rm B\kern-.05em{\sc i\kern-.025em b}\kern-.08em
    T\kern-.1667em\lower.7ex\hbox{E}\kern-.125emX}}
\begin{document}

\title{Extended Short- and Long-Range Mesh Learning\\ for Fast and Generalised Garment Simulation}

\author{
    \IEEEauthorblockN{
        Aoran Liu\IEEEauthorrefmark{2}\textsuperscript{,1}, 
        Kun Hu\IEEEauthorrefmark{3}\textsuperscript{,2,*}\thanks{* Corresponding author.}, 
        Clinton Mo\IEEEauthorrefmark{2}\textsuperscript{,3}, 
        Changyang Li\IEEEauthorrefmark{4}\textsuperscript{,4} and 
        Zhiyong Wang\IEEEauthorrefmark{2}\textsuperscript{,5}
    }
    \IEEEauthorblockA{
        \IEEEauthorrefmark{2}School of Computer Science, 
        The University of Sydney, Darlington, NSW, Australia\\
    }    
    \IEEEauthorblockA{
        \IEEEauthorrefmark{3}School of Science, Edith Cowan University, Joondalup, WA, Australia
    }
    \IEEEauthorblockA{
        \IEEEauthorrefmark{4}Sydney Polytechnic Institute Pty Ltd, Haymarket, NSW, Australia\\
    }
    \IEEEauthorblockA{
        \textsuperscript{1}aliu4429@uni.sydney.edu.au;
        \textsuperscript{2}k.hu@ecu.edu.au;
        \textsuperscript{3}clmo6615@uni.sydney.edu.au;\\
        \textsuperscript{4}chris@ruddergroup.com.au;
        \textsuperscript{5}zhiyong.wang@sydney.edu.au
    }
}

\maketitle

\begin{abstract}

3D garment simulation is a critical component for producing cloth-based graphics. 
Recent advancements in graph neural networks (GNNs) offer a promising approach for efficient garment simulation. However, GNNs require extensive message-passing to propagate information such as physical forces and maintain contact awareness across the entire garment mesh, which becomes computationally inefficient at higher resolutions. To address this, we devise a novel GNN-based mesh learning framework with two key components to extend the message-passing range with minimal overhead, namely the \textit{Laplacian-Smoothed Dual Message-Passing} (LSDMP) and the \textit{Geodesic Self-Attention} (GSA) modules. 
LSDMP enhances message-passing with a Laplacian features smoothing process, which efficiently propagates the impact of each vertex to nearby vertices.
Concurrently, GSA introduces geodesic distance embeddings to represent the spatial relationship between vertices and utilises attention mechanisms to capture global mesh information. The two modules operate in parallel to ensure both short- and long-range mesh modelling.
Extensive experiments demonstrate the state-of-the-art performance of our method, requiring fewer layers and lower inference latency. \footnote{The code is available at https://github.com/adam-lau709/ESLR-Sim.}

\end{abstract}


\section{Introduction}
Garment simulation plays a critical role in various multimedia applications, including 3D animations, video games, fashion design, and virtual try-on experiences. Classical approaches rely on mass-spring systems \cite{terzopoulos1987elastically}, which approximate the forces acting on the garment and distribute them across the vertices of the corresponding 3D garment mesh. However, the commonly used iterative optimisation-based methods for modelling such forces are computationally expensive, posing challenges for various scenarios such as real-time simulation.

In recent years, deep learning networks have emerged as potential alternatives for approximating garment behaviours with lower latency than classical approaches~\cite{wang2023multi}. Graph neural networks (GNNs) in particular have demonstrated remarkable capabilities in modelling complex physical systems, including garment simulation \cite{pfaff2020learning, sanchez2020learning}. 
Nevertheless, in the context of garment simulation, GNNs do not inherently address the iterative optimisation process, as their performances largely depend on their message-passing range, i.e. the geodesic distance over which vertex-wise information can propagate.
To achieve high-quality garment behaviours with proper collision and elasticity handling, a sufficiently large message-passing range is essential. However, this often requires numerous GNN layers, which can be computationally inefficient. 
Therefore, a primary objective in GNN-based methods is to expand each vertex's influence range while minimising the number of message-passing layers. 
For example, recent studies have explored hierarchical graph structures, which enable parallel message-passing at multiple scales \cite{grigorev2023hood, fortunato2022multiscale,wang_huamin_2021}.
Although such methods introduce long-range connections between vertices, they rely on predefined decimation strategies and require the cloth mesh to be reducible to lower resolutions, which is not always feasible. 
This heavily limits their utility when a variety of meshes are necessary, such as in fashion applications.

\begin{figure}
    \centering
    \resizebox{\columnwidth}{!}{%
    \includegraphics[]{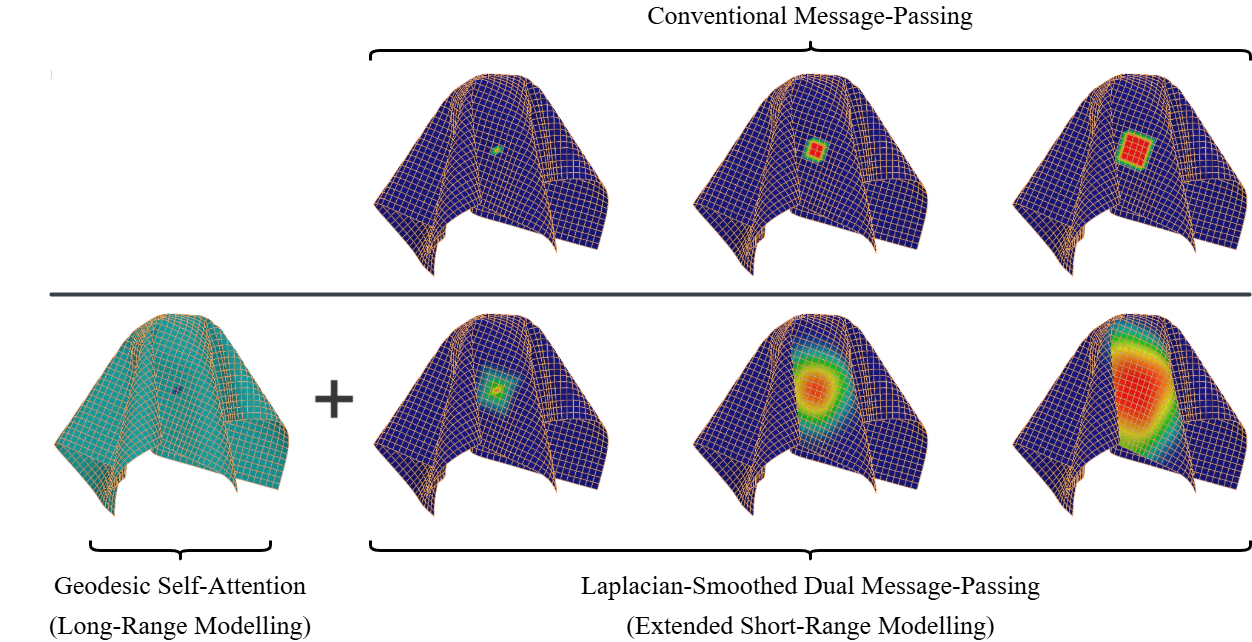}}
    \caption{
    Laplacian smoothing-based propagation allows features to reach distant vertices much earlier in an attenuated fashion, and is vastly more efficient than conventional message-passing. Geodesic self-attention further enables graph-aware long-range connections without message-passing mechanisms.
    }
    \label{fig:MAMP}
\end{figure}

In this paper, we propose a novel GNN-based mesh learning framework to enhance the message-passing range in garment simulation without requiring multi-scale mesh definitions. Our framework introduces two key modules as shown in Fig.~\ref{fig:MAMP}: the \textit{Laplacian-Smoothed Dual Message-Passing} (LSDMP) module for extended short-range modelling, and the \textit{Geodesic Self-Attention} (GSA) module to facilitate long-range vertex connections. 
The LSDMP module employs a dual message-passing design that facilitates message exchange over a wider range. Specifically, each LSDMP layer extends conventional message-passing with a Laplacian-smoothing process for adjacent features aggregation, which efficiently propagates the impact of each vertex to nearby vertices. Complementing the short-range LSDMP module, the long-range GSA module captures global mesh information through attention mechanisms. GSA introduces geodesic distance embeddings to effectively represent the spatial relationships between vertices, enabling the module to focus on long-range vertex interactions.

With the proposed modules, our framework effectively captures both short- and long-range mesh patterns of garments. Comprehensive experiments demonstrate that our method adheres closely to physical laws, while enhancing parameter efficiency and inference latency compared to existing approaches.

In summary, our key contributions are as follows:
\begin{itemize}
    \item We propose a novel GNN-based architecture for garment simulation, leveraging short- and long-range connections.
    \item We devise a \textit{Laplacian-Smoothed Dual Message-Passing} module to extend message-passing of short-range garment features with adjacent vertex-based smoothing. 
    \item We devise a \textit{Geodesic Self-Attention} module to capture long-range  features, with geodesic distance embeddings for vertex connections complementary to LSDMP.
    \item Comprehensive experiments demonstrate our method's superior performance against existing methods, using fewer layers and lower computational latency.
\end{itemize}

\section{Related Work}
\subsection{Pose-Conditioned 3D Garment Simulations}

Conventional 3D garment simulation methods estimate deformations based on real-world fabric physics. Early vertex-based systems used mass-spring models between adjacent vertices to emulate cloth elasticity \cite{terzopoulos1987elastically}, laying the groundwork for later physics-based algorithms \cite{baraff1998large, muller2007position}. While these methods yield realistic results, their high computational cost remains a significant barrier for various applications.

To address this, learning-based methods have emerged, often mapping body pose parameters to garment states \cite{santesteban_2019, Santesteban_2021}, supervised via simulations or 3D scans \cite{saito2021scanimate, ma2020learning}. This approach improves inference speed due to lightweight input data, but ensuring physical accuracy is difficult due to the chaotic nature of garment dynamics. Additionally, generating ground truth simulation data is costly. To overcome this, self-supervised methods optimize garment behavior via physical laws rather than relying on precomputed deformations \cite{bertiche_2021, Santesteban_2022_CVPR}, enabling more robust, physically plausible results.

\subsection{Vertex-Level Garment Learning with Graph Structures}
While pose conditions offer broad guidance for garment dynamics, pose-driven methods heavily rely on learned patterns to capture garment structure, often requiring dedicated models per mesh, limiting flexibility. To address this, graph structures and GNNs have been used to model dynamics with explicit mesh information \cite{pfaff2020learning, fortunato2022multiscale, grigorev2023hood, grigorev2024contourcraft}. These methods predict deformations by learning local vertex interactions via message-passing, enabling generalisation to unseen garments using a unified model.

However, realistic behavior demands a large message-passing range, and increasing message-passing layers can result in prohibitive computational costs. Improving message-passing efficiency is thus key to scalable graph learning. While importance sampling reduces costs \cite{chen2018fastgcn, hamilton2017inductive, chiang2019cluster}, it’s unsuitable for garment simulation where each vertex needs full neighborhood context. Pre-computing message-passing \cite{zhang2022graph, wu2019simplifying} is also suboptimal due to static features. In this study, we aim to explore receptive field expansion techniques for garment simulation, with consideration for dynamic graph configurations and information completeness.

\begin{figure*}[h!]
    \centering
    \includegraphics[width=\textwidth]{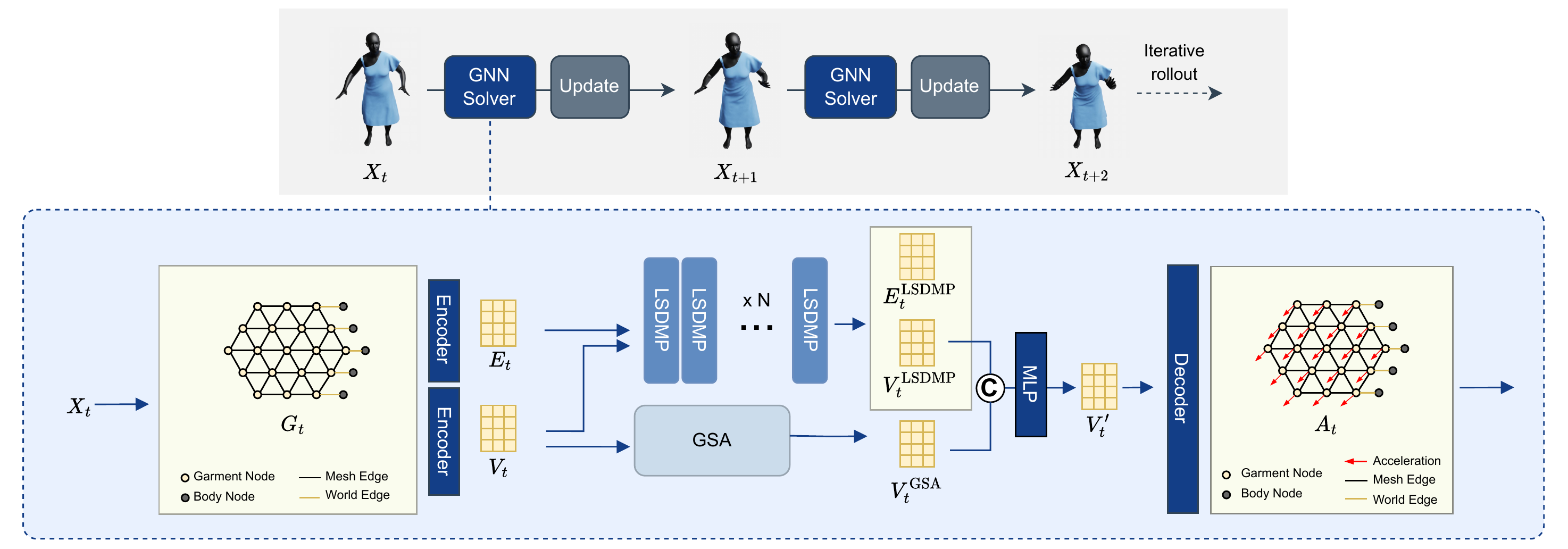}
    \caption{Overview of the proposed method for 3D garment simulation. It consists of two novel modules: LSDMP (Laplacian Smoothed Dual Message-Passing) and GSA (Geodesic Self-Attention), to process extended short and long-range mesh learning, respectively.}
    \label{fig:structure_overview}
\end{figure*}

\section{Methodology}

Our method follows a three-stage pipeline consisting of an encoder, a message-passing module, and a decoder \cite{pfaff2020learning}. We primarily enhance the message-passing module with two novel components: a \textit{Laplacian Smoothed Dual Message-Passing} module and a \textit{Geodesic Self-Attention} module. 
In this section, we provide an overview of this pipeline, followed by a detailed explanation of the proposed components.

\subsection{Garment Simulation Pipeline}

Consider a garment mesh $M_G$ with $n_G$ vertices and a human body mesh $M_B$ with $n_B$ vertices in their undeformed states. Garment simulation can be modeled as an iterative process that estimates a sequence of deformed mesh states $X = \{X_0, X_1, ... , X_T\}$. At each timestep $t$, the state $X_t$ is represented as $X_t=\{X^G_t, X^B_t\}$, where $X^G_t \in \mathbb{R}^{n_G \times 3}$ and $X^B_t \in \mathbb{R}^{n_B \times 3}$ represent the garment and human body mesh states, respectively, each containing their vertex coordinates. 

As illustrated in Fig. \ref{fig:structure_overview}, given the current state $X_{t}$, garment simulation estimates the vertex-wise acceleration of the garment mesh $A_t$, and then uses it to predict the next state of the garment $\hat{X}^G_{t+1}$ as an estimation of $X^G_{t+1}$.  
Particularly, $X_{t}$ is viewed as a graph with its mesh vertices and edges. 
Similarly to MeshGraphNet~\cite{pfaff2020learning}, we add world edges between the body and the garment vertices to model collisions. 
The world edges are based on spatial proximity: if a body and garment vertex pair is within a given distance threshold, a world edge is added between them.

The vertices and edges of $X_{t}$ are encoded into latent features $V_t$ and $E_t$, respectively, with two individual encoders. 
The latent features are then fed into the LSDMP module to perform message propagation. 
In parallel, the GSA module operates on the latent vertex features for a global understanding in terms of the garment. 
The two modules produce updated vertex features $V_t^\text{LSDMP}$ and $V_t^\text{GSA}$. These features are concatenated and processed by an MLP to yield the refined latent vertex features $V_t^\prime$. 
Finally, the decoder uses $V_t^\prime$ to estimate the vertex acceleration $A_t$.
An Euler integration is applied to update the per-vertex velocity of the garment $Q^G_{t+1}= Q^G_{t} + A_{t} \Delta t$, and subsequently, the overall graph for the timestep $t+1$ is obtained by: $\hat{X}^G_{t+1} = X^G_{t} + Q^G_{t+1}\Delta t$, where $\Delta t$ is the associated timestep size. The linear blend skinning result of the garment on the initial body pose is obtained as the initial garment state $X_{0}^{G}$. 

\subsection{Laplacian Smoothed Dual Message-Passing}

\begin{figure}[]
    \centering
    \resizebox{\columnwidth}{!}{%
    \includegraphics[]{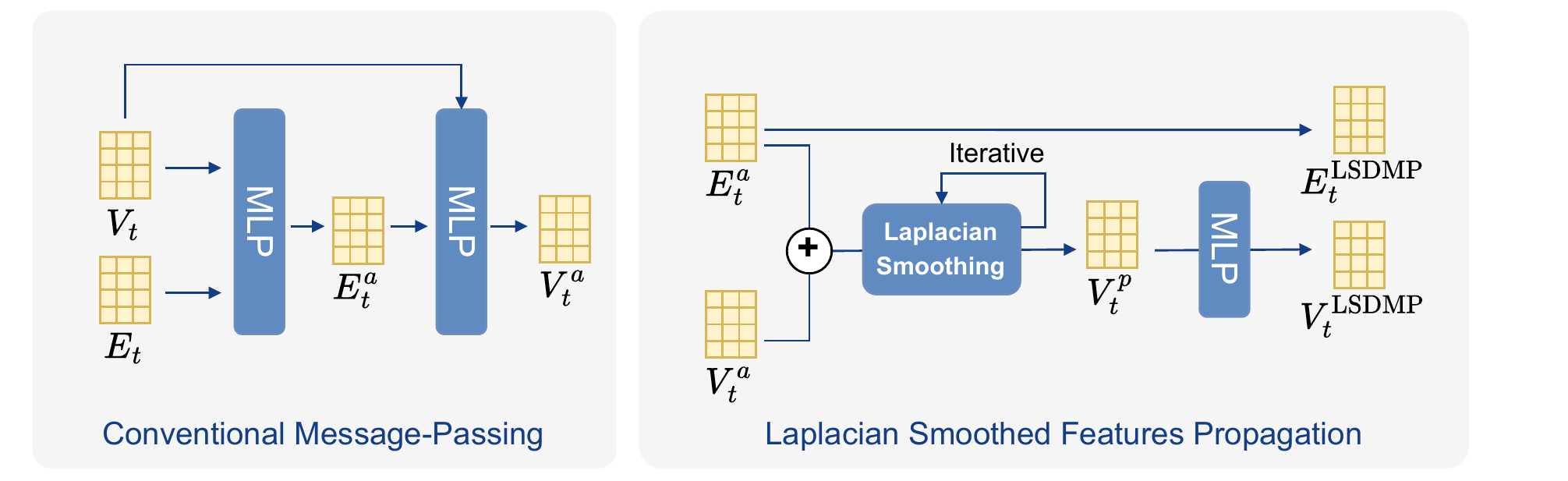}
    }
    \caption{LSDMP extends conventional message-passing using Laplacian-smoothing to efficiently and emblematically propagate vertex features.}
    \label{fig:LSDMP_method}
\end{figure}

The Laplacian Smoothed Dual Message-Passing (LSDMP) module takes the edge features $E_t$ and vertex features $V_t$ as input and outputs the updated features $V_t^\text{LSDMP}$ and $E_t^\text{LSDMP}$, respectively. The LSDMP module consists of a sequential of LSDMP layers. As shown in Fig. \ref{fig:LSDMP_method}, the operations in each layer can be divided into two stages: a conventional message-passing stage~\cite{pfaff2020learning} and a Laplacian features propagation stage.

\begin{figure}[]
    \centering
    \resizebox{\linewidth}{!}{%
    \includegraphics[]{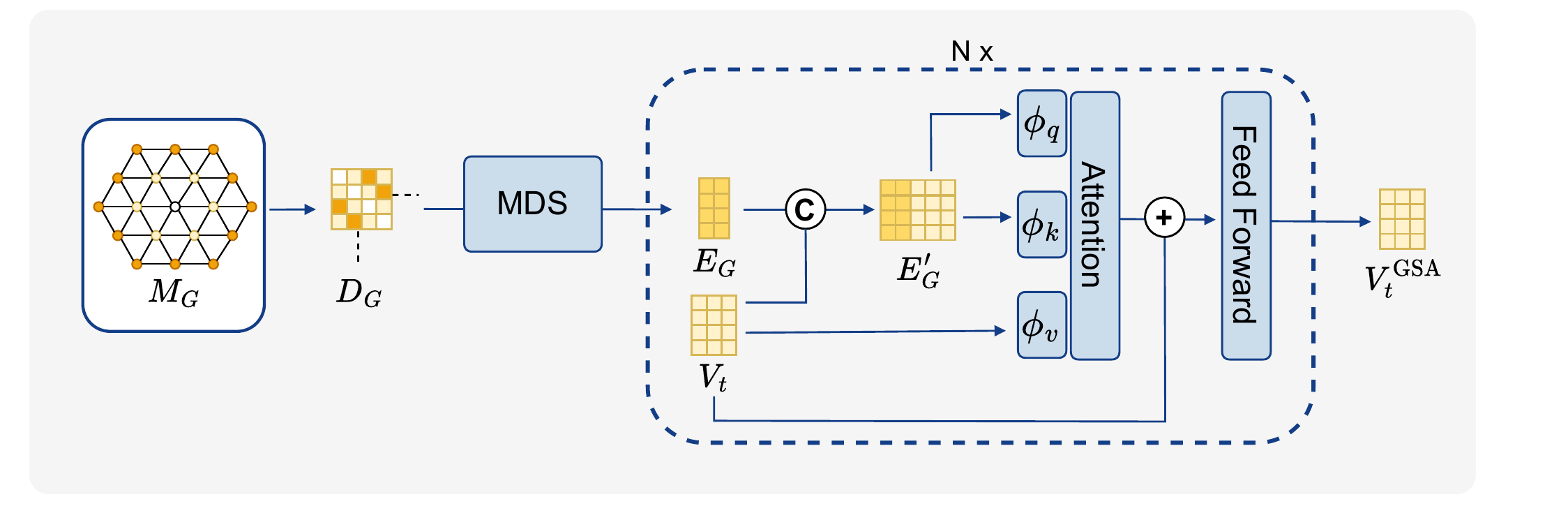}
    }
    \caption{GSA injects geodesic information into vertex features by appending MDS-reduced coordinates derived from the geodesic distance matrix.} 
    \label{fig:GSA}
\end{figure}

In the message-passing stage, each edge first aggregates features from its two connected vertices, followed by an MLP-based update. 
For the edge connecting the $i^\text{th}$ and $j^\text{th}$ vertices, the edge features $e_{ij}^a$ can be obtained through:
\begin{equation}
e_{ij}^{a} = f_e(e_{ij}, v_i, v_j),
\end{equation}
where $f_e$ represents the MLP for the edge-wise update, and $v_i$ and $v_j$ are the corresponding latent vertex features. Next, each vertex compiles information from its connected edges. 
For the $i^\text{th}$ vertex, its compiled edge features $v_i^a$ are computed as: 
\begin{equation}
v_i^{a} = f_v(v_i, \sum_{j}e^{a,m}_{ij}, \sum_{j}e^{a,w}_{ij}),
\end{equation}
where $e^{a,m}_{ij}$ is the features of the connected mesh edges of the $i^{th}$ vertex and $e^{a,w}_{ij}$ represents the features of the connected world edges. 
$f_v$ is an MLP for the vertex-wise update. 

In the Laplacian features propagation stage, we propagate vertex features to nearby vertices using iterative smoothing operations. Laplacian smoothing is computationally efficient, allowing receptive fields to expand with minimal latency. In each iteration, a vertex aggregates the features from its neighbouring vertices and the connected edges.
For the $i^{th}$ vertex, the aggregated features for $v_i^p$ are inferred as: 
\begin{equation}
v_i^{p} = \frac{1}{|\mathcal{N}(i)|} \sum_{j \in \mathcal{N}(i)} (v_j^{a} + e^a_{ij}),
\end{equation}
where $\mathcal{N}(i)$ is the index set for the vertices and edges connected with the $i^{th}$ vertex. The set of aggregated features for all vertices at time $t$ is denoted as 
$V^p_t = \{ v_i^{p} \mid i \in \mathcal{V} \}$,
where \( \mathcal{V} \) represents the set of all vertex indices.
Finally, the vertex features are obtained through an MLP $f'_v$: 
\begin{equation}
V^\text{LSDMP}_t = f_v'(V^p_t).
\end{equation}

\subsection{Self-Attention with Geodesic Distance Embedding}

The LSDMP module enables each garment vertex to gather extended short-range information from its neighboring vertices; however, capturing long-range dependencies and global context across the entire mesh is essential for producing high-quality garment patterns.  
To address this, the GSA module, based on self-attention, is designed to enable each vertex to efficiently capture global information as illustrated in Fig. \ref{fig:GSA}. 

For the self-attention mechanism to function effectively between any two vertices, each vertex must have an understanding of its geodesic relationship with the others.
For this purpose, we introduce geodesic distance embeddings into the self-attention.
Notably, the quadratic complexity of the standard self-attention mechanism is computationally expensive for garments due to the high vertex counts of high-resolution meshes. Inspired by \cite{rampavsek2022recipe}, we adopt the linear attention approach \cite{choromanski2020rethinking} to reduce computational overhead.
However, this approach does not support pairwise information; instead, a vertex-wise geodesic embedding is required.

For the garment mesh, we pre-compute the geodesic distance matrix $D_G \in \mathbb{R}^{n_G \times n_G}$ of the garment, which contains the pairwise geodesic distances between vertices. To formulate $D_G$ in a vertex-wise manner, we generate the per-vertex $k$-dimensional geodesic distance embeddings $E_G \in \mathbb{R}^{n_G \times k}$ by applying Multidimensional Scaling (MDS)\cite{abdi2007metric}. MDS minimises the following objective:
\begin{equation}
f(E_G) = \sum_{i \neq j} \left( d_{ij} - \| d_{i} - d_{j} \|_2 \right)^2,
\end{equation}
where $d_{i}$, $d_{j}$ are the generated geodesic embeddings of the $i^{th}$ and $j^{th}$ vertices derived from $E_G$. $d_{ij}$ represents the geodesic distance between the $i^{th}$ and $j^{th}$ vertices as given by $D_G$. This optimisation problem ensures that the pairwise Euclidean distances in the embedding space closely approximate the pairwise geodesic distances.
By incorporating geodesic embeddings into the queries and keys, the attention module becomes aware of the geodesic relationships between vertices. The GSA module can be stacked multiple times, with the resulting features denoted as $V^\text{GSA}$.



\subsection{Optimisation}

We follow an existing optimisation strategy \cite{grigorev2023hood}, which uses physical losses for unsupervised learning. These losses include minimising \textit{stretch energy} and \textit{collision}-wise penetrations, as well as adhering to \textit{bending stiffness}, \textit{gravity}, \textit{inertia}, \textit{friction} behaviours. Following existing practices, we further adopt these objectives for evaluation purposes.


\section{Experiments}
\subsection{Dataset}

For garment simulation experimentation, we employed the AMASS \cite{AMASS:ICCV:2019} dataset with its human motion for training and evaluation. In total, 56 motion sequences were selected, containing 6,465 frames. 4 of these sequences were used for evaluation and the rest were used for training. 
We also used the same garment template set as \cite{grigorev2023hood} for training, including a T-shirt, tank top, long-sleeved shirt, shorts, long pants and a dress. Additionally, for out-of-domain testing, we employed knee-length pants, a short-sleeve shirt, and a tight dress. 

\subsection{Implementation Details}

Our model consists of separate stacks of LSDMP and GSA layers. The LSDMP module consists of 15 stacked layers, with each layer conducting a conventional message-passing step followed by 3 Laplacian smoothing steps. The GSA module is designed to be lightweight, comprising 4 stacked attention blocks with one single-head self-attention layer.
We trained the model over 100,000 iterations as in \cite{grigorev2023hood}, on an RTX 4070 Ti 12GB GPU, completing the process in approximately 16 hours.  
The training process involves randomly sampling a body pose from a motion sequence and a garment template from the training template set to use as input for each iteration.
We compute the linear blend skinning result of the garment for this pose, which serves as the initial garment state \cite{SMPL-X:2019, grigorev2023hood}. 

\subsection{Simulation Performance on Native Garments}
 
To evaluate the effectiveness of our method, we provide a comparison to state-of-the-art methods including SSCH\cite{Santesteban_2021}, MSN\cite{liu2023material}, MAT\cite{Li2024NeuralGarmentDynamics}, MeshGraphNet\cite{pfaff2020learning} and HOOD\cite{grigorev2023hood} regarding physical errors with the trained garment meshes on novel poses. 
SSCH and MSN are pose-conditioned and garment-specific methods, requiring separate models to be trained for each garment. In contrast, MAT, MeshGraphNet and HOOD are generalised methods capable of simulating diverse garments with a single unified model. 
Note that all GNN-based methods, including our LSDMP module, are configured with 15 message passing layers.

Table \ref{table:quantitative} shows that our method achieves the lowest total loss among all evaluated approaches, demonstrating strong physical awareness. Specifically, our method shows a significant advantage in stretch energy, a critical metric for maintaining realistic deformation and simulation stability. A lower stretch energy indicates effective constraint modelling, which contributes to accurate and stable garment simulations. This improvement stems from our two proposed modules: the LSDMP module expands the message-passing range, enabling better modeling of elastic behaviors, while the GSA module leverages geodesic embeddings to capture global acceleration distributions, leading to more stable simulations.


In addition to stretch energy, our method also achieves significant improvement on collision handling, due to the integration of the GSA module. By incorporating geodesic embeddings, vertices can instantly detect collisions in geodesically close regions and respond without the latency or attenuation typically introduced by message-passing.
The ablation study in Table \ref{table:ablation} affirms these observations. The model with both modules achieve the lowest physics loss, indicating the contributions of the two modules in producing physically accurate simulation results.


\begin{table}[]
\caption{Physical simulation performance on native garments}
\resizebox{\linewidth}{!}{
\setlength{\tabcolsep}{0.2em}
\begin{tabular}{|l|c|c|c|c|c|c|c|}
\hline
Method        & Stretch$\downarrow$ & Bending$\downarrow$ & Inertia$\downarrow$ & Collision$\downarrow$ & Friction$\downarrow$ & Gravity$\downarrow$ & Total$\downarrow$  \\ \hline
SSCH          & 2.09E+00            & 7.20E-03            & 9.33E-03            & 8.15E-02              & 3.14E-03             & -2.15E-01           & 1.97E+00           \\
MSN           & 1.92E-01            & 8.17E-03            & 5.50E-03            & 6.26E-02              & 2.82E-03             & -2.37E-01           & 3.39E-02           \\
MAT           & 1.07E+00            & 9.41E-03            & \textbf{4.50E-03}   & 2.36E-02              & 2.84E-03             & -1.59E-01           & 9.54E-01           \\
MeshGraphNet  & 2.77E-01            & 6.81E-03            & 5.94E-03            & 6.08E-02              & 2.22E-03             & \textbf{-3.06E-01}  & 4.68E-02           \\
HOOD          & {\ul 1.42E-01}      & {\ul 5.34E-03}      & 5.20E-03            & {\ul 1.96E-02}        & {\ul 2.03E-03}       & {\ul -2.84E-01}     & {\ul -1.10E-01}    \\ \hline
\textbf{Ours} & \textbf{8.69E-02}   & \textbf{4.80E-03}   & {\ul 4.82E-03}      & \textbf{1.10E-02}     & \textbf{1.79E-03}    & -2.49E-01           & \textbf{-1.40E-01} \\ \hline
\end{tabular}
}
\label{table:quantitative}
\end{table}


\begin{table}[]
\caption{Ablation study on LSDMP and GSA modules}
\resizebox{\columnwidth}{!}{%
\setlength{\tabcolsep}{0.2em}
\begin{tabular}{|l|c|c|c|c|c|c|c|}
\hline
Method      & Stretch$\downarrow$ & Bending$\downarrow$ & Inertia$\downarrow$ & Collision$\downarrow$ & Friction$\downarrow$ & Gravity$\downarrow$ & Total$\downarrow$  \\ \hline
Baseline    & 2.77E-01            & 6.81E-03            & 5.94E-03            & 6.08E-02              & 2.22E-03             & \textbf{-3.06E-01}  & 4.68E-02           \\
LSDMP       & {\ul 9.31E-02}      & {\ul 5.29E-03}      & 5.09E-03            & 1.49E-02              & {\ul 1.87E-03}       & {\ul -2.53E-01}     & {\ul -1.33E-01}    \\
GSA         & 1.22E-01            & 6.67E-03            & \textbf{4.58E-03}   & {\ul 1.37E-02}        & 1.88E-03             & -2.51E-01           & -1.02E-01          \\
LSDMP + GSA & \textbf{8.69E-02}   & \textbf{4.80E-03}   & {\ul 4.82E-03}      & \textbf{1.10E-02}     & \textbf{1.79E-03}    & -2.49E-01           & \textbf{-1.40E-01} \\ \hline
\end{tabular}
}
\label{table:ablation}
\end{table}

\begin{table}
\caption{Physical simulation performance on unseen garments}
\resizebox{\columnwidth}{!}{
\setlength{\tabcolsep}{0.2em}
\begin{tabular}{|lccccccc|}
\hline
\multicolumn{1}{|l|}{Method}        & \multicolumn{1}{c|}{Stretch$\downarrow$}           & \multicolumn{1}{c|}{Bending$\downarrow$}           & \multicolumn{1}{c|}{Inertia$\downarrow$}           & \multicolumn{1}{c|}{Collision$\downarrow$}         & \multicolumn{1}{c|}{Friction$\downarrow$}          & \multicolumn{1}{c|}{Gravity$\downarrow$}            & Total$\downarrow$                         \\ \hline
\multicolumn{8}{|c|}{Knee-length Pants}                                                                                                                                                                                                                                                                                          \\ \hline
\multicolumn{1}{|l|}{MeshGraphNet} & \multicolumn{1}{c|}{4.52E-01}          & \multicolumn{1}{c|}{4.05E-03}          & \multicolumn{1}{c|}{8.03E-03}          & \multicolumn{1}{c|}{{\ul 5.50E+00}}    & \multicolumn{1}{c|}{\textbf{2.55E-03}} & \multicolumn{1}{c|}{\textbf{-8.50E-01}} & {\ul 5.12E+00}                \\
\multicolumn{1}{|l|}{HOOD}         & \multicolumn{1}{c|}{4.97E-01}          & \multicolumn{1}{c|}{\textbf{3.85E-03}} & \multicolumn{1}{c|}{{\ul 7.83E-03}}    & \multicolumn{1}{c|}{5.50E+00}          & \multicolumn{1}{c|}{2.71E-03}          & \multicolumn{1}{c|}{{\ul -8.49E-01}}    & 5.16E+00                      \\
\multicolumn{1}{|l|}{Ours}         & \multicolumn{1}{c|}{\textbf{1.69E-01}} & \multicolumn{1}{c|}{{\ul 3.92E-03}}    & \multicolumn{1}{c|}{\textbf{7.75E-03}} & \multicolumn{1}{c|}{\textbf{5.41E+00}} & \multicolumn{1}{c|}{{\ul 2.59E-03}}    & \multicolumn{1}{c|}{-8.28E-01}          & \textbf{4.76E+00}             \\ \hline
\multicolumn{8}{|c|}{Tight Dress}                                                                                                                                                                                                                                                                                         \\ \hline
\multicolumn{1}{|l|}{MeshGraphNet} & \multicolumn{1}{c|}{1.27E-01}          & \multicolumn{1}{c|}{{\ul 6.18E-03}}    & \multicolumn{1}{c|}{{\ul 5.28E-03}}    & \multicolumn{1}{c|}{3.19E-03}          & \multicolumn{1}{c|}{{\ul 2.40E-03}}    & \multicolumn{1}{c|}{\textbf{-2.31E-01}} & -8.67E-02                     \\
\multicolumn{1}{|l|}{HOOD}         & \multicolumn{1}{c|}{{\ul 1.13E-01}}    & \multicolumn{1}{c|}{6.20E-03}          & \multicolumn{1}{c|}{5.86E-03}          & \multicolumn{1}{c|}{\textbf{1.45E-03}} & \multicolumn{1}{c|}{2.61E-03}          & \multicolumn{1}{c|}{{\ul -2.22E-01}}    & {\ul -9.28E-02}               \\
\multicolumn{1}{|l|}{Ours}         & \multicolumn{1}{c|}{\textbf{7.43E-02}} & \multicolumn{1}{c|}{\textbf{5.47E-03}} & \multicolumn{1}{c|}{\textbf{4.89E-03}} & \multicolumn{1}{c|}{{\ul 1.51E-03}}    & \multicolumn{1}{c|}{\textbf{2.26E-03}} & \multicolumn{1}{c|}{-1.98E-01}          & \textbf{-1.10E-01}            \\ \hline
\multicolumn{8}{|c|}{Short-sleeved Shirt}                                                                                                                                                                                                                                                                                         \\ \hline
\multicolumn{1}{|l|}{MeshGraphNet} & \multicolumn{1}{c|}{7.02E-02}          & \multicolumn{1}{c|}{3.92E-03}          & \multicolumn{1}{c|}{3.06E-03}          & \multicolumn{1}{c|}{1.53E-03}          & \multicolumn{1}{c|}{{\ul 1.23E-03}}    & \multicolumn{1}{c|}{\textbf{6.95E-02}}  & \multicolumn{1}{l|}{1.49E-01} \\
\multicolumn{1}{|l|}{HOOD}         & \multicolumn{1}{c|}{{\ul 4.94E-02}}    & \multicolumn{1}{c|}{{\ul 3.84E-03}}    & \multicolumn{1}{c|}{{\ul 2.94E-03}}    & \multicolumn{1}{c|}{\textbf{7.46E-04}} & \multicolumn{1}{c|}{1.23E-03}          & \multicolumn{1}{c|}{{\ul 7.62E-02}}     & {\ul 1.34E-01}                \\
\multicolumn{1}{|l|}{Ours}         & \multicolumn{1}{c|}{\textbf{3.44E-02}} & \multicolumn{1}{c|}{\textbf{3.52E-03}} & \multicolumn{1}{c|}{\textbf{2.82E-03}} & \multicolumn{1}{c|}{{\ul 1.24E-03}}    & \multicolumn{1}{c|}{\textbf{1.12E-03}} & \multicolumn{1}{c|}{8.39E-02}           & \textbf{1.27E-01}             \\ \hline
\end{tabular}
}
\label{table:generalizability}
\end{table}

\begin{figure}
    \centering
    \resizebox{0.9\columnwidth}{!}{%
    \includegraphics[]{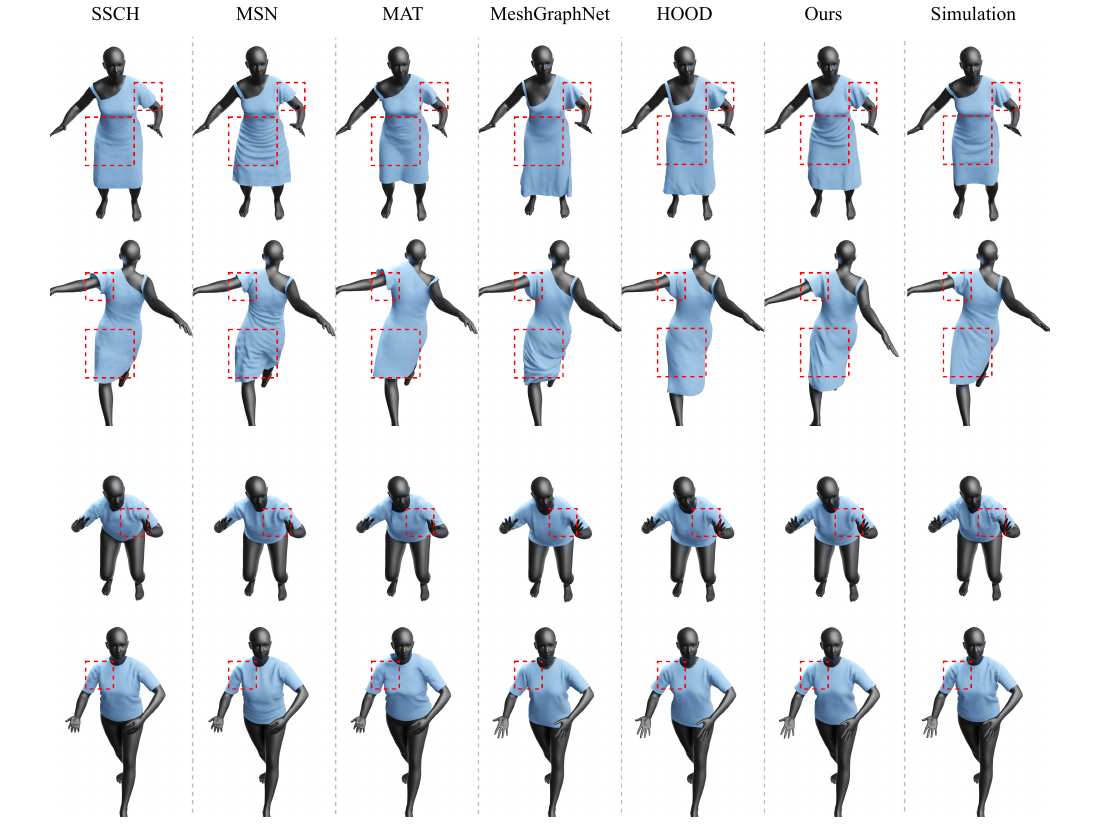}
    }
    \caption{Qualitative comparison with SOTA methods. We further include a physical simulation method - ARCSim \cite{narain2012adaptive} for reference.}
    \label{fig:qualitative}
\end{figure}

We present the qualitative examples in Fig.~\ref{fig:qualitative}, which demonstrate that our method produces more realistic dynamics and finer wrinkle details in more accurate regions compared to other methods. 
In terms of garment dynamics, pose-conditioned models like SSCH and MSN tend to cling too tightly to the body and overreact to movement, resulting in stiff, unnatural garment behavior. In contrast, our graph-based method models vertex interactions directly, leading to more realistic dynamics.
In terms of detail fineness, the results from SSCH, MAT, and HOOD appear overly smoothed, lacking fine-grained wrinkle details, whereas MeshGraphNet and MSN predict such details excessively.
Our method finds a strong middle ground with wrinkle intensity, and more importantly, tends to keep wrinkle details contained exclusively within garment regions that should realistically have wrinkles. 

\begin{figure}
    \centering
    \resizebox{0.9\columnwidth}{!}{%
    \includegraphics[]{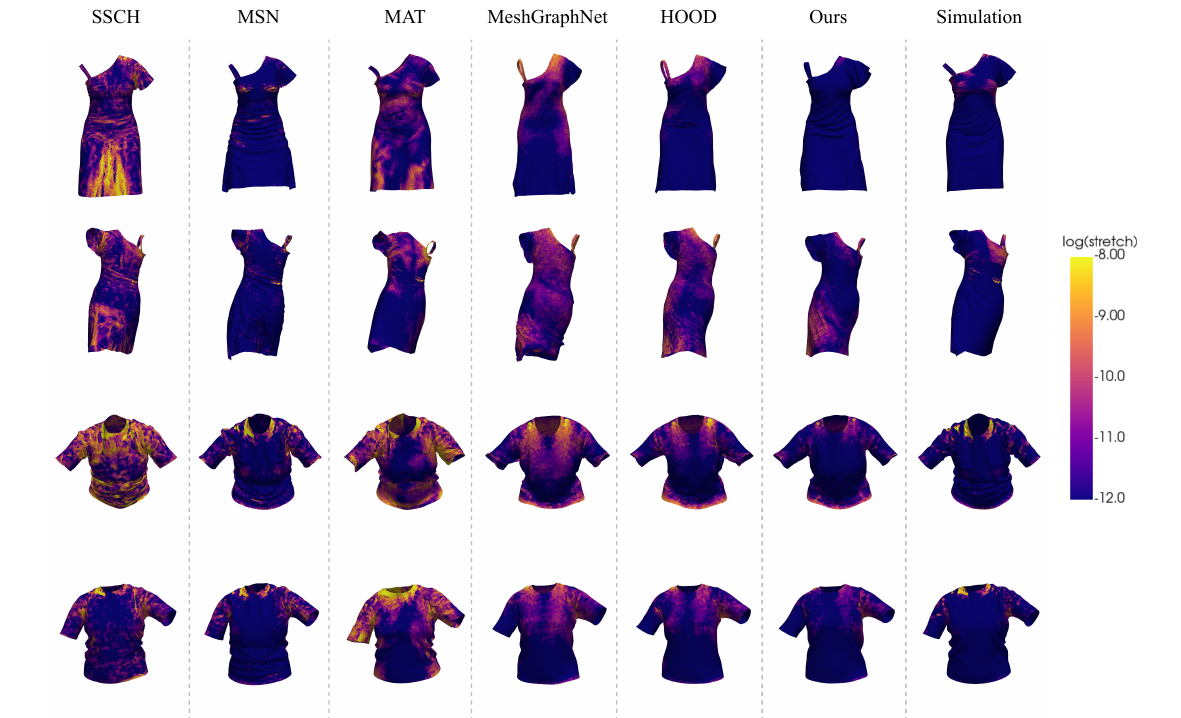}
    }
    \caption{Stretch energy distribution.}
    \label{fig:energy}
\end{figure}

Fig.~\ref{fig:energy} visualizes the stretch energy distribution for selected frames, showing that our method consistently predicts lower stretch energy than others, indicating more physically stable results. It should be noted, however, that body motion can reduce energy entropy, which explains the high stretch energy in collision-prone areas like collars, even in physics-based methods.

\subsection{Generalisability to Unseen Garments and Avatars}



We evaluate our method’s adaptability to arbitrary garment meshes using three unseen garments: a tight dress, a short-sleeved top, and knee-length pants. Accuracy is compared against MeshGraphNet and HOOD, both supporting mesh-agnostic simulation. As shown in Table \ref{table:generalizability}, our method consistently achieves the lowest total physics loss. However, our method's improvements in collision handling are nullified with unseen meshes, likely due to the incompatibility of geodesic distance matrices between different meshes for GSA.



\begin{figure}[] \centering
    \resizebox{0.9\columnwidth}{!}{
    \includegraphics[]{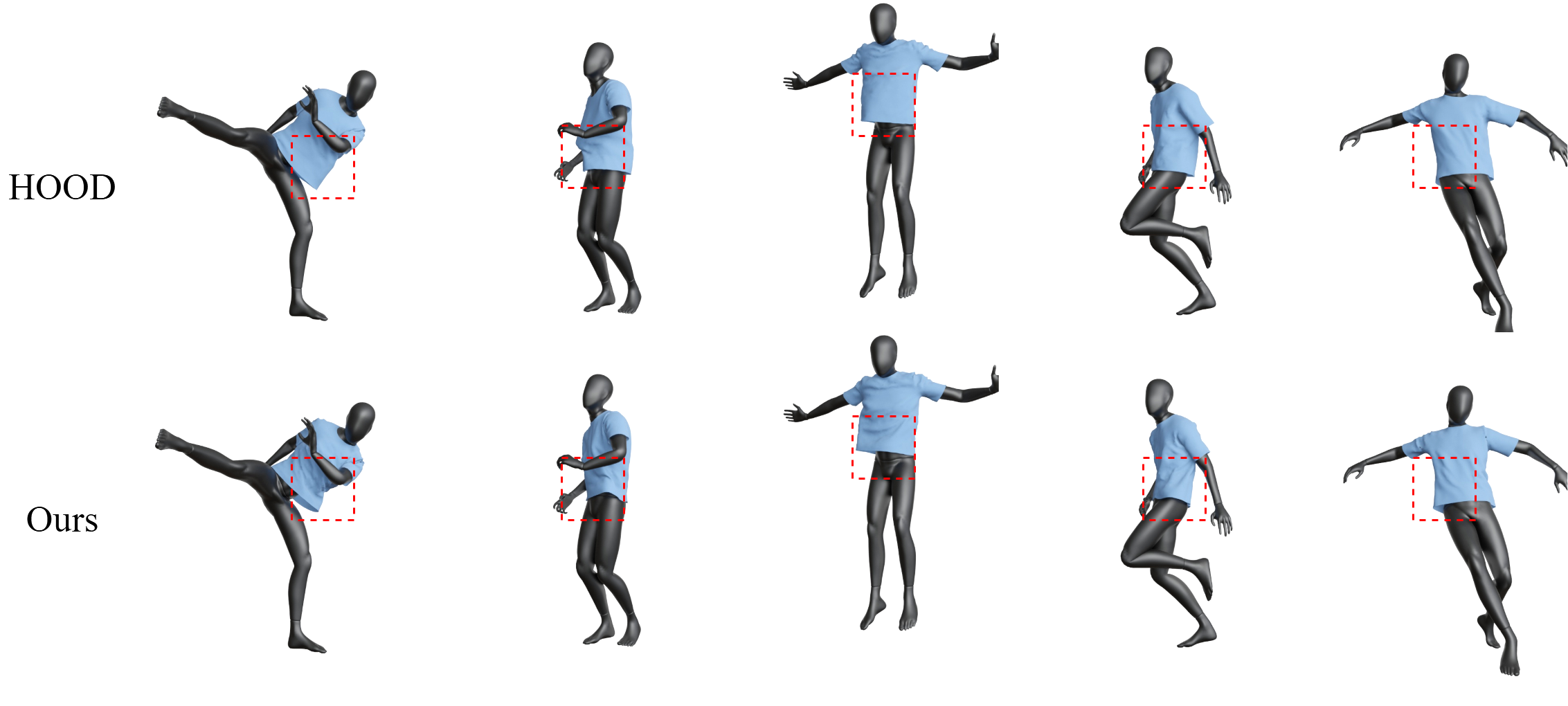}
    }
    \caption{Qualitative results with MIXAMO sequences.} \label{fig:mixamo}
\end{figure}

To assess generalisability to new avatars, we further present evaluation results on unseen MIXAMO\footnote{https://www.mixamo.com} motion sequences with a new avatar and garment in Fig. \ref{fig:mixamo} and Table \ref{table:mixamo}. Our method outperforms the SOTA method HOOD, demonstrating strong generalisability.


\begin{table}[]
\centering
\caption{Quantitative results with MIXAMO sequences} \label{table:mixamo}
\resizebox{\columnwidth}{!}{
\begin{tabular}{|c|ccccccc|}
\hline
Model & Stretch$\downarrow$ & Bending$\downarrow$ & Inertia$\downarrow$ & Collision$\downarrow$ & Friction$\downarrow$ & Gravity$\downarrow$ & Total$\downarrow$ \\ \hline
HOOD  & 1.74E-01            & 1.78E-02            & 3.29E+00            & 7.98E-03              & \textbf{1.42E-03}    & 6.67E-03            & 3.49E+00          \\
Ours  & \textbf{1.27E-01}   & \textbf{1.48E-02}   & \textbf{2.82E+00}   & \textbf{5.69E-03}     & 1.54E-02             & \textbf{5.43E-03}   & \textbf{2.98E+00} \\ \hline
\end{tabular}
}
\end{table}

\subsection{Model Efficiency and Inference Latency}

To demonstrate the layer/parameter efficiency of our method, we compare it against two baseline models based on the conventional message-passing proposed in MeshGraphNet \cite{pfaff2020learning}. The baseline models are configured with 15 and 48 layers, whereas our method uses 10 and 15-layer setups. As shown in Table \ref{table:param-efficiency}, both setups of our method significantly outperform the 15-layer baseline in terms of total physics loss, while achieving performance comparable to the 48-layer baseline. Furthermore, our 15-layer setup uses only $\sim40\%$ of the parameters needed by the 48-layer baseline while delivering comparable results with half the inference latency and lower memory requirements, highlighting its parameter efficiency.

\begin{table}[]
\caption{Inference latency against Conventional Message Passing}
\resizebox{\columnwidth}{!}{
\setlength{\tabcolsep}{0.2em}
\begin{tabular}{|c|c|c|c|c|c|}
\hline
Model                         & \# of Layers & \# of Parameters & Total Loss$\downarrow$ & Latency (ms)$\downarrow$ & Memory Usage (MB)$\downarrow$ \\ \hline
\multirow{2}{*}{Baseline} & 48           & 12.0M            & -1.66E-01  & 133.8        & 320               \\
                              & 15           & 3.85M            & 4.68E-02   & 57.8         & 290               \\ \hline
\multirow{2}{*}{Ours}         & 15           & 4.75M            & -1.40E-01  & 68.8         & 292               \\
                              & 10           & 3.23M            & -1.21E-01  & 54.0           & 286               \\ \hline
\end{tabular}
}
\label{table:param-efficiency}
\end{table}

\section{Limitation \& Future Work}


While the proposed method improves GNN-based garment simulation efficiency, GNNs still incur high computational costs at ultra-high mesh resolutions, limiting real-time applicability. A promising future direction is applying super-resolution to low-resolution outputs instead of simulating high-resolution meshes directly. Additionally, while our GSA module reduces self-collision artifacts, no proper optimisation measures are currently in place for self-collision, leaving the issue partially unresolved. Future work could explore incorporating a self-collision objective \cite{grigorev2024contourcraft} or using implicit representation\cite{Sun2024-cj} to further address this problem.

\section{Conclusion}

In this paper, we present a GNN-based garment simulation framework that enhances existing methods by incorporating two new modules: LSDMP and GSA. These modules extend the GNN's message-passing range, enabling the capture of both extended short- and long-range features with minimal computational overhead. Experiments have demonstrated that the self-attention and attenuation features of our method achieve superior simulation quality with a compact model size and low inference latency.

\bibliographystyle{IEEEbib}
\bibliography{icme2025references}
\end{document}